# Evolutionary Design of Numerical Methods: Generating Finite Difference and Integration Schemes by Differential Evolution

C.D. Erdbrink[1,2,3], V.V. Krzhizhanovskaya[1,2], P.M.A. Sloot[1,2,4]
1. National Research University of Information Technologies, Mechanics and Optics, Saint Petersburg, Russia
2. University of Amsterdam, The Netherlands
3. Deltares, The Netherlands
4. Nanyang Technological University, Singapore

**Abstract**
Classical and new numerical schemes are generated using evolutionary computing. Differential Evolution is used to find the coefficients of finite difference approximations of function derivatives, and of single and multi-step integration methods. The coefficients are reverse engineered based on samples from a target function and its derivative used for training. The Runge-Kutta schemes are trained using the order condition equations.
An appealing feature of the evolutionary method is the low number of model parameters. The population size, termination criterion and number of training points are determined in a sensitivity analysis. Computational results show good agreement between evolved and analytical coefficients. In particular, a new fifth-order Runge-Kutta scheme is computed which adheres to the order conditions with a sum of absolute errors of order $10^{-14}$. Execution of the evolved schemes proved the intended orders of accuracy. The outcome of this study is valuable for future developments in the design of complex numerical methods that are out of reach by conventional means.

## Introduction

In this paper coefficients of classical numerical schemes for differentiation and integration are generated by an evolutionary algorithm (EA). The widespread use and therefore importance of finite difference methods and Runge-Kutta methods is self-evident. Basic finite difference schemes have straightforward analytical derivations, but complex computational science problems often require the design of more sophisticated methods lacking standard approaches. The novel perspective investigated in this study embraces the application of EAs for designing numerical formulae.

Runge-Kutta methods comprise a family of integration schemes for solving initial value problems of Ordinary Differential Equations (ODEs), the most famous being the fourth-order RK4 Runge-Kutta method. For orders past two the systems of order condition equations are underdetermined, this hampers analytical derivation. Tsitouras & Famelis (2012) note that traditional heuristics based on computing Jacobians are less suitable for finding Runge-Kutta schemes than EAs, since determining the derivatives of the variables as they appear in the order conditions is cumbersome (Rothlauf 2011).

In the last couple of decades several classes of EAs have been developed and applied to an abundance of fields in science and engineering (Eiben & Smith 2013). The power of evolutionary computing is increasingly reflected in its ability to tackle more general computational problems that are analytically intractable and, moreover, inaccessible to conventional numerical methods. In a seminal double pendulum experiment Schmidt & Lipson (2009) showed the effectiveness and versatility of symbolic regression based on Genetic Programming (GP) by deriving classical physical laws from experimental data without using a priori knowledge about underlying physical characteristics.

This concept of contriving universal equations from elementary building blocks without human intervention can also be applied to numerical mathematics itself. EAs generally become interesting when approximate solutions are acceptable and small improvements over existing solutions are extremely valuable (Poli et al. 2007). In addition, it is required that testing of the quality of candidate solutions should be unambiguous. The problem at hand, finding discrete computational schemes for estimating derivatives and integrals, meets these conditions: inexact schemes are acceptable as the schemes themselves represent approximations, and the performance of a scheme is easily checked by applying it to a function with known properties.

For the modelling task of generating both a scheme's structure as well as its coefficients, the parse tree method of GP provides an appropriate approach. However, if the equation structure is given and the goal is to find only the coefficients, it becomes an optimization problem and other EAs can be used. Differential Evolution (DE) is a relatively new heuristic technique (Storn & Price 1997) for global optimization. Since its birth, several improvements were introduced (Das & Suganthan 2011) and the power of DE is reflected by outperforming a number of stochastic optimization algorithms such as Adaptive Simulated Annealing (Storn & Price 1997) and Particle Swarm Optimization (Vesterstrøm & Thomsen 2004); it beats Genetic Algorithms (GA) on many benchmarks as well (Tušar & Filipič 2007, Hegerty et al. 2009).

Previously, the application of EAs to evolve computational algorithms was mostly aimed at solving domain-specific problems. For example, Spector et al. (1998) used GP to find quantum computing algorithms with superior performance. A number of past studies utilize EAs in more general contexts to solve systems of equations or somehow tune traditional numerical methods to improve performance. For example, He et al. (2000) propose a hybrid algorithm that combines the classical Successive Over-Relaxation method (SOR) for solving linear systems of equations with an evolutionary technique to evolve the relaxation factor.

A study by BaniHani (2007) uses GA to determine points and weights for an integration procedure within a meshfree method for solving boundary value problems. To the best of our knowledge, Martino & Nicosia (2012), Tsitouras & Famelis (2012) and Nanayakkara et al. (1999) are the only efforts so far to apply an evolutionary algorithm to derive Runge-Kutta-related methods. The first study advocates the use of EAs in numerical analysis from the viewpoint of algebraic geometry, but does not present concrete values of coefficients, nor does it discuss accuracy. The second study considers neural networks and DE to find Runge-Kutta-Nyström pairs for solving a second order ODE problem occurring in astronomy. The third study uses an EA with the purpose of finding shape parameters of Runge-Kutta-Gill neural networks, through radial basis function networks, for the identification of robot arm dynamics.

In the field of finite difference methods, an effort by Haeri & Kim (2013) uses GA to optimize boundary characteristics of compact finite difference equations. These recent studies illustrate the relevance of applying EAs to find involved numerical recipes and provide the motivation for our research, where we apply DE for deriving general, i.e. domain-independent numerical schemes.

The aim of our study is to apply Differential Evolution to find coefficient sets of classical numerical schemes for approximating derivatives and integrals. This will not only produce new coefficients for schemes where fully analytical derivation of coefficients is unattainable, but the reverse-engineering process will also yield insight into how to apply the evolutionary heuristic and what the resulting accuracy is. This paves the way for the application of EAs to more complex numerical stencils of general use in applied mathematics and engineering.

The next section treats related work in past research. Then a background on the studied numerical schemes is given and it is explained how the coefficients will be represented. This is followed by a description of the computational method used in this study. Then the modelling results are presented; this section consists of a sensitivity analysis and presentation and discussion of training and validation results. Lastly, conclusions are drawn and a short outlook is given on possibilities for future work.

**Background**
1. Finite difference derivative estimates
Common finite difference formulae are of the type central, forward or backward. Table 1 gives two examples, viz. the second-order central formula and the second-order forward formula for estimating the first-order derivative.

*Table 1. Representation of finite difference schemes.*

| finite difference formula | vector representation |
|---|---|
| $f'(x) \approx \dfrac{f(x+h) - f(x-h)}{2h}$ | $(1/2 \quad 0 \quad -1/2)^T$ |
| $f'(x) \approx \dfrac{-3f(x) + 4f(x+h) - f(x+2h)}{2h}$ | $(-3/2 \quad 2 \quad -1/2)^T$ |

Here $h$ is the step size. By putting all coefficients in the numerator (thus leaving only $h$ in the denominator) each approximation scheme is defined by its coefficients $m_i \in \mathbb{Q}$ for all terms $m_i \cdot f(x + n_i h)$ with $n_i \in \mathbb{Z}$. The number of terms grows with increasing order of accuracy and the coefficients can, among other ways, be determined via the Taylor series. Note additionally that the same coefficients used in a forward scheme can also be used in an equivalent backward scheme with symmetric terms $f(x \pm h)$. For even order derivatives, the coefficients are equal for corresponding terms, but for odd order derivatives the coefficients are multiplied by -1.

A central scheme of order $p$ contains $p$ nonzero terms, where p is even. The $p$ coefficients of a central scheme will be searched for by evolving vectors of sizes $p$ and $p+1$. The formula's skeleton assumed for runs with vectors of size $p$ consists of terms $f(x \pm ph)$ with $q$ = 1, 2,.., $p/2$ and vectors of size $p+1$ contain the extra term $f(x)$. A forward scheme of order $p$ contains $p+1$ nonzero terms and will be represented only by vectors of size $p+1$. For the forward scheme the skeleton consists of the terms $f(x + qh)$ with $q$ = 0, 1,.., $p$.

2. Runge-Kutta schemes

Runge-Kutta schemes for solving initial value problems are summarized in Butcher tableaus or arrays (Butcher 2008). Table 2 shows this notation and also gives the corresponding vector representation that is used in our study.

*Table 2. Representation of explicit Runge-Kutta schemes.*

| | Butcher tableau | | | | | vector representation of explicit scheme |
|---|---|---|---|---|---|---|
| $c_1$ | $a_{11}$ | $a_{12}$ | … | … | $a_{1s}$ | |
| $c_2$ | $a_{21}$ | $a_{22}$ | … | … | $a_{2s}$ | |
| ⋮ | ⋮ | ⋮ | ⋮ | ⋮ | ⋮ | $(a_{21}\ a_{31}\ a_{32}\ a_{41}\ \ldots\ a_{s1}..a_{ss-1}\ w_1 \ldots w_s)^T$ |
| $c_s$ | $a_{s1}$ | $a_{s2}$ | … | $a_{ss\text{-}1}$ | $a_{ss}$ | |
| | $w_1$ | $w_2$ | … | … | $w_s$ | |

Where $c_i$ are called the nodes, $w_i$ are the weights and $s$ is the stage. The Butcher tableau holds the coefficients for the formula

$$y_n = y_{n-1} + \sum_{i=1}^{s} w_i k_i$$

with

$$k_i = hf\left(t_n + c_i h; y_n + \sum_{j=1}^{s} a_{ij} k_j\right)$$

For explicit schemes the matrix $A$ containing entries $a_{ij}$ is lower triangular, i.e. $a_{ij}$ = 0 for i ≤ j. Furthermore, the consistency condition requires that the sum of the row elements in $A$ is equal to the node at that level: $a_{i1} + \cdots + a_{is} = c_i, \forall i$ (Hairer et al. 1993). Assuming consistency therefore implies that explicit schemes can be encoded by the vector given in Table 2. For instance, a 4-stage scheme is represented by the vector $(a_{21}\ a_{31}\ a_{32}\ a_{41}\ a_{42}\ a_{43}\ w_1\ w_2\ w_3\ w_4)^T$.

The stage $s$ of a scheme fixes its maximum order of accuracy. Order condition equations determine if an order is actually achieved (Butcher 2008). For example, a 2-stage scheme ($s$ = 2) has second order accuracy if $w_1 + w_2$ = 1 and $w_2 a_{21}$ = ½. The order conditions up to order five are included in Appendix I. Note that the 2-stage scheme has easy closure: choose a value for $a_{21}$ and the other coefficients are fixed. By contrast, higher order schemes are far more demanding; 6-stage schemes need to meet 17 order conditions to attain order 5. This study considers 3-, 4- and 6-stage schemes, reaching a maximum order of 5.

3. Adams-Bashforth schemes

The Adams-Bashforth method is an explicit linear multi-step integration method (Butcher 2008). Depending on the target order of accuracy, a certain number of preceding prediction values are used in the new estimate. The general formula for a $k$-term method reads

$$y_n = y_{n-1} + h \sum_{i=1}^{k} \beta_i f(t_{n-i}, y_{n-i})$$

The maximum achievable accuracy order of a *k*-term method is *k*. The order conditions can be used to find analytical values of the coefficients. The coefficients representation is straightforward for the Adams-Bashforth schemes: each vector contains all $\beta_i$ for $1 \leq i \leq k$, for computing a scheme of intended order *k*.

**Method for generating computational schemes by an evolutionary algorithm**
The original DE algorithm, called DE/rand/1/bin (Storn & Price 1997), assumes constant control parameters *CR* for crossover and scaling factor *F* for mutation. With the advent of dynamic parameters adjustment (Eiben et al. 1999), later versions of DE have control parameters that are varied as part of the evolution (Brest et al. 2006). Our study is based on DE/rand/1/bin and adopts the self-adaptive method by Choi et al. (2013) where each individual carries its own pair of control parameters. Updates take place every generation after the selection by taking the average values of the control parameters of individuals that evolved successfully in the past generation and then adding variation to these averages by drawing from a Cauchy distribution. A pseudo-code of the algorithm used in the present study is added in Appendix II.

The initial population consists of floating-point vectors with uniformly random entries between -1 and 1, but no value constraints are imposed during the evolution. To ensure a healthy balance between population diversity and exploitation, in every generation five randomly chosen individuals (other than the highest scoring individual) are replaced by completely new individuals. The algorithm has four key model parameters: $CR_0$, $F_0$, population size *NP* and the number of computed generations, where the zero indices denote the initial values of the control parameters applied to all individuals. Because of the random nature of the self-adaptation, these initial values have negligible influence on the result of the evolution; this study uses $CR_0$ = 0.25 and $F_0$ = 0.6. The population size and number of computed generations will be determined in a sensitivity study.

1. Finite difference derivative estimates
Training consists of comparing first order derivative estimates calculated by the candidate formula with analytical values sampled from a target derivative function. A bell-shaped curve and its derivative are used as target function, see Figure 1 and Appendix III.

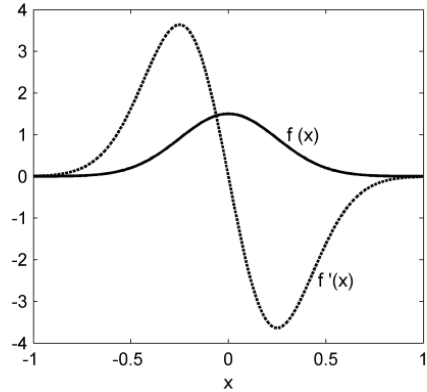

*Figure 1. Function pair used for training coefficients of finite difference schemes and Adams-Bashforth schemes*

The fitness function measures the absolute errors between candidate values and analytical coefficient values:

$$F_{FD} = \log \sum_{k=1}^{n} |v_k - v_k^*|$$

Where $\boldsymbol{v} \in \mathbb{R}^n$ is the vector containing the theoretical values, $\boldsymbol{v}^* \in \mathbb{R}^n$ is a candidate vector and $n = \dim_\mathbb{R}(\mathbb{R}^n)$. The logarithm is applied to deal with the large range of values. In our paper the evolution seeks to minimize the fitness functions.

2. Runge-Kutta schemes
The candidate coefficient vectors are evaluated by substitution into the order condition equations. The fitness of the candidate Runge-Kutta schemes is therefore defined by

$$F_{RK} = \log \sum_{k=1}^{\#C} |C_k - C_k^*|$$

where $C_k$ is the analytical value of order condition equation $k$; these are the right-hand-side values in Appendix I. $C_k^*$ is the candidate order conditions vector computed by inserting all necessary entries of the candidate coefficients vector into the order condition equations. #$C$ is the number of order conditions.

3. Adams-Bashforth schemes
Fitness evaluation is defined identically to the finite difference schemes, but now the candidate scheme is evaluated by integrating $f'$ ($f$ as defined in Figure 1 and Appendix III) and the result is compared with the analytical values sampled from $f$. The necessary starting points are generated by a Runge-Kutta scheme of the same order of the Adams-Bashforth scheme that is aimed for.

**Results and discussion**

*Sensitivity analysis*
Before starting the training process of the finite difference runs the impact of the number of generations and the population size are considered, as well as the characteristics of the target function. The target function pair was chosen in such a way that the derivative has both positive and negative values. Preliminary runs for the sixth order central difference scheme showed that the exact shape (periodicity, number of extremes, symmetry, etc.) of target function $f$ has no notable influence on accuracy, as long as $f'$ is continuous and has a reasonable variation such that the fitness can discriminate between good and bad approximations. Clearly, the function pair $f(x) = 4x + 7$, $f'(x) = 4$ would be unsuitable.

Varying the range of function $f$ (Figure 1) between 0.1 and 1000 for an equal sampling step size $h$ and equal number of training points, giving a variation in the range of $f'$ between 0.4 and 4850, yielded no differences in achieved accuracy. Next, the step size was varied while keeping the total number of training points the same. This also had no noticeable impact on accuracy. Figure 2 shows the results when the reverse is done: varying the number of training points while keeping $h$ constant, for three different values of $h$.

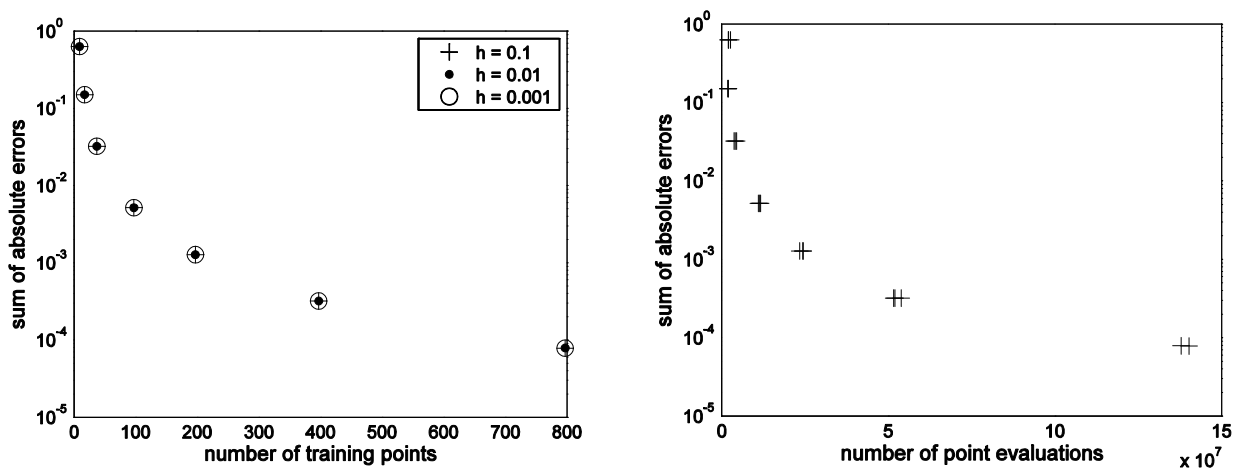

*Figure 2. Sensitivity to training data for the central sixth order approximation of a first order derivative. Left: varying number of training points in the target function while keeping sample step size* h *constant, for three different step sizes. Right: sum of absolute errors as function of the number of point evaluations. The two plots are based on the same 21 runs with a population size of 150. On the vertical axes the sum of absolute errors over all coefficients is plotted on a log-scale.*

Figure 2 (left) shows that the number of training points, i.e. the number of unique points sampled from the target function, is a major factor for accuracy. Additionally, it underlines the fact that the step size at which the target function is sampled has no influence at all. The results also show that the improvements in accuracy are getting smaller as we increase the number of training points ($N$): the sum of errors gets 1000 times lower with $N$ increasing from 1 to 200; and only 10 times lower with $N$ increasing from 200 to 800.
The number of point evaluations is the product of the number of computed generations, population size and number of calls to training points. The sum of absolute errors as a function of number of point evaluations (Figure 2 right) displays a similar relation as for the dependence on the number of training points, which shows that using more training points does not imply significantly more generations needed to attain convergence. Comparing the two plots of Figure 2, we conclude that runs with a similar number of training points and a different step size achieve the same accuracy (left plot) and need only a slightly different number of generations for convergence, resulting in small variations in the number of point evaluations per run (right plot).

The computation time of a run depends linearly on the number of point evaluations and therefore also approximately linearly on the number of training points. The plots in Figure 2 suggest that improving the accuracy by including more than 800 training points comes with disproportionally higher computational costs. The plotted trial run for the sixth order central scheme with 800 training points consisted of about $1.4 \cdot 10^8$ point evaluations and resulted in a sum of absolute errors of $7.8 \cdot 10^{-5}$.

The impact of using different population sizes is investigated by making 25 test runs for the finite difference schemes (based on 200 training points and varying between 50 and 1000 individuals) and for the Runge-Kutta schemes (using the 6-stage scheme and varying between 50 and 500 individuals). The results are plotted in Figure 3.

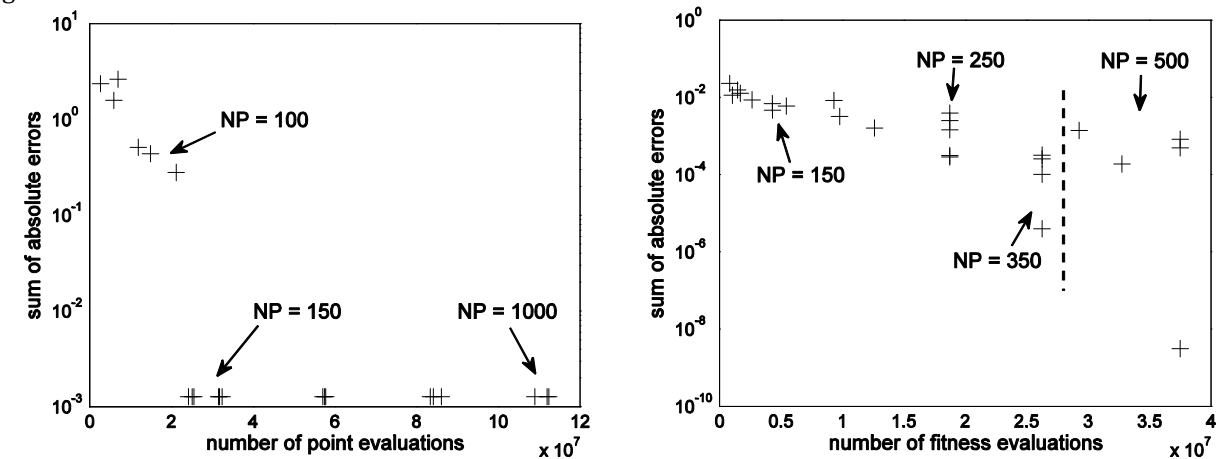

*Figure 3. Sensitivity to population size. Left: finite difference sixth-order central approximation of first order derivative, changing population size NP from 50 to 1000 and keeping number of training points constant at 200. Right: 6-stage Runge-Kutta runs, varying population size NP between 50 and 500. The sum of absolute errors is plotted vertically on log-scale.*

The sensitivity to the population size is quite obvious for the finite difference schemes (Figure 3 left): population sizes of 150 or more outperform all smaller population sizes. For population sizes smaller than or equal to 100, the sum of absolute training errors is significant – these schemes never give accurate derivative approximations. Improvements by using populations beyond 150 are marginal compared to the increase in computational costs. Averaged over three runs per population size, the training error for a population of 1000 is less than 0.1% smaller than for a population of 150, but the computation time is a factor 4.4 higher. Using a population of 150 is therefore optimal for the finite difference runs.

The population size has a different effect on the evolution of Runge-Kutta schemes (Figure 3 right). While winning individuals become exponentially better with higher population size, the standard deviation of the attained accuracies also increases with number of fitness evaluations. This means that more runs are required to improve accuracy and at the same time that each run becomes more expensive. Based on this sensitivity, a population size of 350 is chosen for the Runge-Kutta schemes, in combination with a high number of runs.

The number of generations that make up one run is regulated by a termination criterion. The evolutionary algorithm stops if the fitness of the best individual has not changed in a pre-set number of consecutive generations or if a maximum number of generations is reached. Both parameters were determined empirically throughout the sensitivity runs by looking at the convergence of the runs. Figure 4 illustrates the working of the termination criterion for the 6-stage Runge-Kutta scheme. The 'example run' never actually converges, as it regularly makes very small improvements but does not escape local optima, it stopped because the maximum number of generations was reached. The 'winning run' experiences a few considerable jumps before converging.

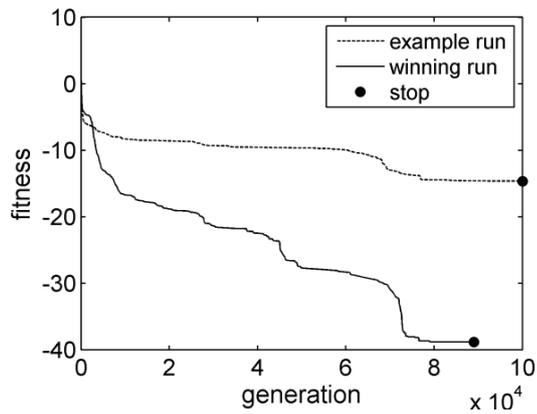

*Figure 4. Evolution of Runge-Kutta 6-stage scheme, showing the fitnesses as a function of generations for a relatively poor example run and for the winning run (out of 100 runs). Termination points are indicated.*

The plot in Figure 4 also clarifies that this particular problem demands a large number of generations. The sensitivity runs for Runge-Kutta were done with a maximum of 75,000 generations and a stop after 7,500 generations without improving the best fitness score. This was extended to respectively 100,000 and 10,000 generations to ensure convergence for the more successful runs. However, for the lower order schemes convergence proved to occur much sooner. Table 3 summarizes the model parameters chosen after the sensitivity study.

*Table 3. Chosen model settings.*

|  | population size $NP$ | stop after this number of generations without improvement | maximum number of generations | number of training points $N$ | number of runs per order |
|---|---|---|---|---|---|
| Finite Difference runs | 150 | 250 | 2,500 | 800 | 10 |
| Adams-Bashforth runs | 150 | 250 | 2,500 | 6400 | 10 |
| Runge-Kutta runs orders ≤ 4 | 350 | 500 | 5,000 | - | 100 |
| Runge-Kutta runs order 5 | 350 | 10,000 | 100,000 | - | 100 |

The sensitivity analysis reveals a clear difference between the computations based on a target function and those based on order conditions. As discussed, for the Runge-Kutta runs the large variation in results and the slow convergence makes it imperative to make many runs with a high number of generations. Conversely, the finite difference and Adams-Bashforth runs have quick convergence and moreover attain very similar optima for different runs, so that a small number of relatively short runs suffice. Compared to the runs of the finite difference schemes, computation of Adams-Bashforth methods requires less point evaluations, so that more training points can be included in a comparable simulation time.

*Training results of derivative schemes*
The error used in the fitness computation is the difference between candidate derivative values and the analytical values of the target function. However, since the coefficients are known from theory it is more sensible to assess the training results by the absolute errors between computed and analytical coefficient values. These errors are plotted in Figure 5 (left). Appendix IV contains complete tables with the resulting coefficients of the best runs for all computed configurations.

Figure 5 compares the results of schemes of different orders. The general trend is that the mean absolute error of a finite difference scheme increases with the order. The forward scheme exhibitsa virtually exponential error growth, whereas the central scheme errors deteriorate only past the sixth order. The central runs where an extra $f(x)$-term was added are only slightly worse than the runs where this was not done; the corresponding coefficients are close to zero (see Appendix IV). Figure 5 (left) also indicates that the two special schemes attain similar accuracy as the traditional schemes. Also, the results of the computed Adams-Bashforth methods are plotted in the same figure. They are relatively accurate and show an exponential trend up to the fourth order.

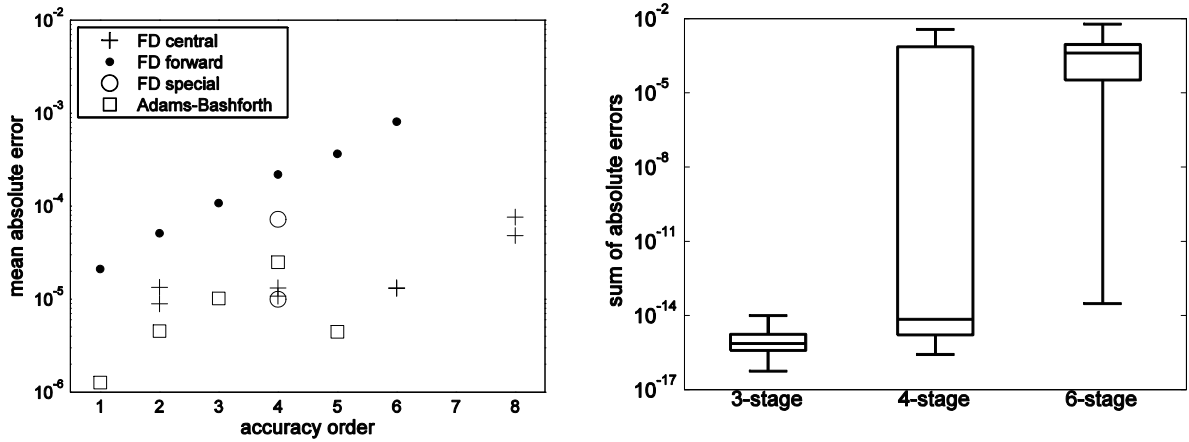

*Figure 5. Left: mean absolute error of evolved coefficients as function of accuracy order for training based on target points of the function in Figure 1. Right: sum of absolute errors for all hundred runs for the computed Runge-Kutta schemes of stages 3, 4 and 6. In the box plots, the central line is the median, the box edges indicate first and third quartiles and the whiskers extend to the most extreme data points.*

*Training results of integration schemes*

The definition of the training error for the Runge-Kutta schemes is based on computed and analytical values of the order conditions. Unlike the analytically tractable finite difference schemes, the computed coefficients for the Runge-Kutta methods do not resemble known values. Box plots are shown in Figure 5 (right) for all the runs performed for stages 3, 4 and 6. The whiskers of the box plots extend to the extreme values. Most interestingly, the best evolved schemes have absolute errors, summed over all coefficients, better than $10^{-13}$ for all stages. The variation in results is largest for the 4-stage scheme, which suggests that in hindsight the termination criterion could have been more conservative. Nevertheless, the accuracy of the evolved schemes for stages 3 and 4 is bounded by rounding-off errors; the used software (MATLAB) allows for 16 digits in the double-precision (64 bit) floating point format. Better results would arguably be possible by using symbolic representation, but this slows down the computations considerably.

*Validation*

In validation tests we applied the best evolved numerical schemes to the differentiation and integration of unseen functions and compared the errors of these numerical solutions (deviations from the analytical solution) to the errors of the classical (analytically derived) numerical schemes. By varying the step size, the slopes of the resulting graphs reveal the orders of accuracy. The Runge-Kutta schemes are applied to the same non-autonomous initial value problem that was used in Boyce and DiPrima (2001) to compute errors; all other methods are applied to a rapidly varying exponential function (see Appendix IV). The validation runs are plotted in Figure 6 for the finite difference schemes and in Figure 7 for the integration schemes. In all cases, the normalized local error is considered at the same *x*-location for all schemes.

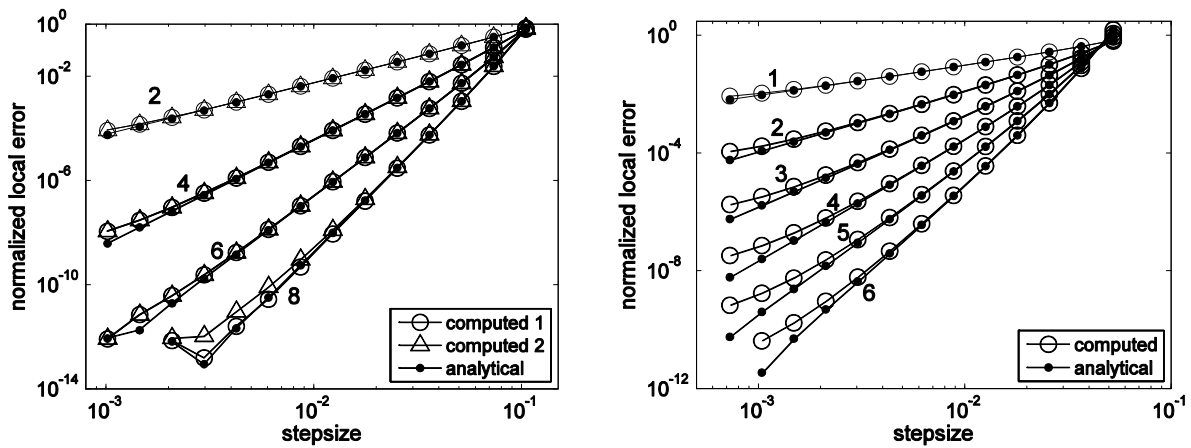

*Figure 6. Validation errors of computed finite difference schemes for approximating first order derivative. Left: central schemes; Right: forward schemes. The orders of accuracy are indicated in the plots.*

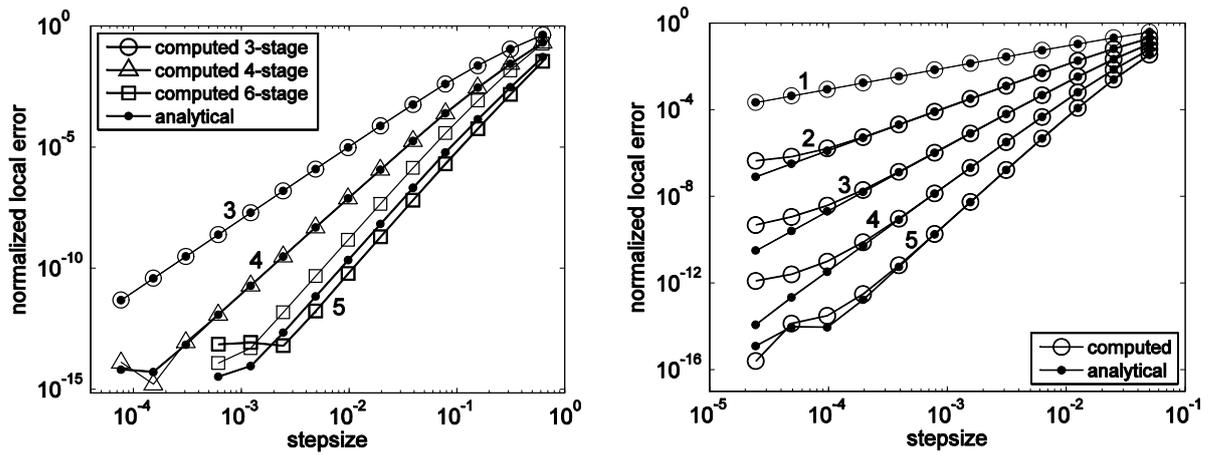

*Figure 7. Validation errors of computed integration schemes compared to results of analytically derived schemes. Left: Runge-Kutta schemes; Right: Adams-Bashforth schemes. The orders of accuracy are indicated in the plots.*

The plots confirm that the intended orders of accuracy are indeed reached. This follows from the fact that there is little deviation between computed and analytical schemes. Also, it was verified that the lines run parallel to the theoretical lines $h^n$, for order $n$. The plots of Figures 6 and 7 exhibit two sources of deviation from the expected straight line: due to rounding-off errors and due to the computed scheme not being as accurate as the analytical scheme. The latter error type is manifested by the bending line ends of the forward scheme (Figure 6, right) and the Adams-Bashforth scheme towards smaller step sizes (Figure 7, right). The software-dependent machine rounding-off error affects both computed and analytical schemes, as can be seen for example for the fourth-order (4-stage) Runge-Kutta scheme around $h = 10^{-4}$ (Figure 7, left). For the Runge-Kutta fifth-order method (6-stage), the results of the *two* best evolved schemes are plotted along with the results of a theoretically derived scheme (Butcher 2008). Remarkably, one of these schemes achieves a bettter accuracy than the theoretical scheme: the lowest line in Figure 7 (left) has the smallest errors. This is a coincidental feature related to the choice of the test function. Several additional tests with different functions (not shown to avoid figure overcrowding) exhibited the same order of accuracy with errors following the same $h^5$ trend), but the errors were not always lower than those of the analytical scheme.

**Conclusion and outlook**

This study has laid a foundation for automatic derivation of discrete numerical schemes by using evolutionary computing. This can be a highly prized approach in situations where analytical solutions are absent and other heuristics fail.

It was shown in this paper how coefficients of widely used numerical methods can be computed up to practicable accuracies using Differential Evolution. This was done by representing coefficient sets as floating-point vectors. Two different training procedures were applied: based on a target function pair $f$ and $f\,'$ and based on order condition equations. Finite difference methods for approximating first derivatives were computed up to order 8 (central schemes) and order 6 (forward schemes); also two non-standard schemes of order 4 were derived. The multi-step Adams-Bashforth integration method was computed up to order 5. Accurate training of the finite difference and Adams-Bashforth schemes, which employed samples from the target function pair, was found to depend primarily on the number of training points and not on sample frequency or target function shape. Moreover, there appeared to be a threshold value for the population size required for attaining reasonable accuracies. As a result, the sum of absolute training errors showed a predictable decaying exponential relation with the number of point evaluations.
Explicit Runge-Kutta methods were trained by promoting adherence to the order conditions, yielding schemes of stages 3, 4 and 6, up to order 5 and with absolute errors summed over all order conditions in the order of $10^{-14}$. The influence of population size proved to be more diffuse than for the target-function training. This made it necessary to have more and longer runs (up to 100,000 generations) to achieve convergence. Furthermore, schemes of higher order, having more terms and coefficients, required larger population size (for order condition training) or inclusion of more training points (for target function training).

The research underlines the effectiveness of the Differential Evolution algorithm as an easy-to-implement heuristic with few model parameters. The accurate results of the Runge-Kutta runs showed its persisting capability to avoid local optima, provided that the termination criterion is appropriately tuned so that runs do not end prematurely.

Although the optimization problem of this study could have been tackled by other means, the evolutionary approach was chosen because it can be naturally combined with Genetic Programming, which can be applied to evolve the structures of various types of equations (e.g., Ryan & Keijzer 2003). This can be thought of as a form of system identification. Future applications could include the computational derivation of numerical stencils for solving partial differential equations. As a starting point, the feasibility of applying Genetic Programming to solving the convection-diffusion equation was recently shown by Howard et al. (2011). In addition, practical research aims could be finding tailor-made solutions for notorious numerical issues arising, for instance, in connection with boundary conditions or model decomposition in mesh-based Computational Fluid Dynamics. Flow models that span different spatial scales such as in Erdbrink et al. (in print) could benefit from these developments.

**Acknowledgements**
is the authors acknowledge the support of the Leading Scientist Program of the Russian Federation under contract 11.G34.31.0019. CDE also acknowledges the support of Deltares.

**Appendix I**

The stage number of an explicit Runge-Kutta scheme has a fixed relation to the maximum achievable order of accuracy and the number of order condition equations required to reach that order. See the table. Butcher (2008) provides details:

*Table 4. Overview of order conditions for explicit Runge-Kutta methods*

| number of stages | 1 | 2 | 3 | 4 | 5 | 6 |
|---|---|---|---|---|---|---|
| number of coefficients | 1 | 3 | 6 | 10 | 15 | 21 |
| max. achievable order | 1 | 2 | 3 | 4 | 4 | 5 |
| number of order conditions | 1 | 2 | 4 | 8 | 8 | 17 |

So, for example, for a 6-stage scheme to have order 5, its coefficients need to satisfy 17 order conditions. These order conditions are listed below in their concise forms. Although for explicit schemes many terms disappear, since $a_{ij} = 0$ for $i \leq j$ and $c_1 = 0$, these systems are not generally solvable without extra assumptions and analysis. The following four order condition equations exist if a three-stage Runge-Kutta scheme is to have order three:

$$\sum_{i=1}^{s} w_i = 1$$

$$\sum_{i=1}^{s} w_i c_i = \frac{1}{2}$$

$$\sum_{i=1}^{s} w_i c_i^2 = \frac{1}{3}$$

$$\sum_{j=1}^{s}\sum_{i=1}^{s} w_i a_{ij} c_j = \frac{1}{6}$$

For a four-stage Runge-Kutta scheme to have order four, not only the above equations should hold, but also the following four order conditions:

$$\sum_{i=1}^{s} w_i c_i^3 = \frac{1}{4}$$

$$\sum_{j=1}^{s}\sum_{i=1}^{s} w_i c_i a_{ij} c_j = \frac{1}{8}$$

$$\sum_{j=1}^{s}\sum_{i=1}^{s} w_i a_{ij} c_j^2 = \frac{1}{12}$$

$$\sum_{k=1}^{s}\sum_{j=1}^{s}\sum_{i=1}^{s} w_i a_{ij} a_{jk} c_k = \frac{1}{24}$$

A five-stage scheme cannot meet the requirements for order six. For a six-stage scheme to have order five, these nine additional order condition equations must hold:

$$\sum_{i=1}^{s} w_i c_i^4 = \frac{1}{5}$$

$$\sum_{j=1}^{s}\sum_{i=1}^{s} w_i c_i^2 a_{ij} c_j = \frac{1}{10}$$

$$\sum_{j=1}^{s}\sum_{i=1}^{s} w_i c_i a_{ij} c_j^2 = \frac{1}{15}$$

$$\sum_{k=1}^{s}\sum_{j=1}^{s}\sum_{i=1}^{s} w_i c_i a_{ij} a_{jk} c_k = \frac{1}{30}$$

$$\sum_{k=1}^{s}\sum_{j=1}^{s}\sum_{i=1}^{s} w_i a_{ij} c_j a_{ik} c_k = \frac{1}{20}$$

$$\sum_{j=1}^{s}\sum_{i=1}^{s} w_i a_{ij} c_j^3 = \frac{1}{20}$$

$$\sum_{k=1}^{s}\sum_{j=1}^{s}\sum_{i=1}^{s} w_i a_{ij} c_j a_{jk} c_k = \frac{1}{40}$$

$$\sum_{k=1}^{s}\sum_{j=1}^{s}\sum_{i=1}^{s} w_i a_{ij} a_{jk} c_k^2 = \frac{1}{60}$$

$$\sum_{l=1}^{s}\sum_{k=1}^{s}\sum_{j=1}^{s}\sum_{i=1}^{s} w_i a_{ij} a_{jk} a_{kl} c_l = \frac{1}{120}$$

**Appendix II**
The pseudo code of the Differential Evolution algorithm:

**1. Initialization**
Set generation counter to $G = 0$
Initialize population of $NP$ vector individuals $X_{1,G}, \ldots, X_{NP,G}$ with $X_{i,G} = [x_{1,i,G}, x_{2,i,G}, \ldots, x_{D,i,G}]$ with $D$ the vector dimension, equal to the number of coefficients to be evolved, and $i = 1, \ldots, NP$
FOR $i = 1$ to $NP$ DO
   FOR $j = 1$ to $D$ DO
     $x_{j,i,0}$ = -1 + rand[0,1]*2 with rand[0,1] uniformly random from [0,1]
   END FOR
   $CR_{i,0} = 0.25$
   $F_{i,0} = 0.6$
END FOR
Initialize adaptation parameters by $CR_{avg,0} = CR_{i,0}$ and $F_{avg,0} = F_{i,0}$

WHILE termination criterion is not satisfied DO
   $G = G + 1$
   $k = 0$
   FOR $i = 1$ to $NP$ DO
     Determine fitness $f(X_{i,G})$ of individual $i$ by running the scheme with coefficients $x_{1,i,G}, x_{2,i,G}, \ldots, x_{D,i,G}$
     **2. Mutation operation**
     Generate a mutant vector $V_{i,G} = [v_{1,i,G}, v_{2,i,G}, \ldots, v_{D,i,G}]$ from three donor vectors $X_{r1}, X_{r2}, X_{r3}$ randomly selected from the set of vector individuals $X_{1,G-1}, \ldots, X_{NP,G-1}$ as:
     $V_{i,G} = X_{r1} + F_{i,G}*(X_{r2} - X_{r3})$

     **3. Crossover**
     Select a random integer $j_{rand}$ from the integer set $[1, \ldots, D]$
     Generate a trial vector $U_{i,G} = [u_{1,i,G}, u_{2,i,G}, \ldots, u_{D,i,G}]$ as follows:
     FOR $j = 1$ to $D$ DO
       IF rand[0,1] $\leq CR_{i,G-1}$ OR $j = j_{rand}$ THEN $u_{j,i,G} = v_{j,i,G}$
       ELSE $u_{j,i,G} = x_{j,i,G-1}$
       END IF
     END FOR

     **4. Selection**
     Determine fitness $f(U_{i,G})$ of trial vector $i$ by running the scheme with coefficients $u_{1,i,G}, u_{2,i,G}, \ldots, u_{D,i,G}$
     IF $f(U_{i,G}) \leq f(X_{i,G-1})$ THEN
       $X_{i,G} = U_{i,G}$
       $f(X_{i,G}) = f(U_{i,G})$
       $k = k + 1$
       $F\_memory_k = F_{i,G-1}$
       $CR\_memory_k = CR_{i,G-1}$
     ELSE
       $X_{i,G} = X_{i,G-1}$
       $f(X_{i,G}) = f(X_{i,G-1})$
     END IF
   END FOR

   **5. Control parameters adaptation**
   IF $k \neq 0$ THEN
     $F_{avg,G}$ = mean(**F_memory**)
     $CR_{avg,G}$ = mean(**CR_memory**)
   END IF
   FOR $i = 1$ to $NP$ DO
     $F_{i,G} = C(0,\gamma_F) + F_{avg,G}$ with $C(0,\gamma_F)$ randomly drawn from the Cauchy distribution with location parameter 0 and half-width at half-maximum value $\gamma_F = 0.1$
     $CR_{i,G} = C(0,\gamma_{CR}) + CR_{avg,G}$ with $C(0,\gamma_{CR})$ randomly drawn from the Cauchy distribution with location parameter 0 and half-width at half-maximum value $\gamma_{CR} = 0.1$
     Truncate $F_{i,G}$ between 0.1 and 1
     Truncate $CR_{i,G}$ between 0 and 1
   END FOR

   **6. Add new individuals**
   Order individuals according to their fitness
   FOR $n = 1$ to 5 DO
     Select random integer $q$ from the integer set $[1, \ldots, NP]$ other than index of fittest individual
     Re-initialize $X_{q,G}$

END FOR

    Determine value of termination criterion
END WHILE

**Appendix III**
Target function pair used for training of Finite Difference schemes (Figure 1):

$$f(x) = 1.5e^{-0.5x^2}$$
$$f'(x) = -1.5xe^{-0.5x^2}$$

for -4 < x < 4, with *h* = 0.01.

Target function pair used for validation of Runge-Kutta schemes (after Boyce & DiPrima, 2001):

$$y' = 1 - x + 4y; \ y(0) = 1, \text{ with solution } y = \frac{1}{4}x - \frac{3}{16} + \frac{19}{16}e^{4x}$$

Target function pair used for validation of Finite Difference and Adams-Bashforth schemes:

$$f(x) = 2e^{18x} \text{ and } f'(x) = 36e^{18x}$$

# Appendix IV

*Table 5. Results of evolving coefficients of central finite difference approximations of first derivative.*

| theoretical order of accuracy | | vector length | f(x-4h) | f(x-3h) | f(x-2h) | f(x-h) | f(x) | f(x+h) | f(x+2h) | f(x+3h) | f(x+4h) | sum of absolute errors |
|---|---|---|---|---|---|---|---|---|---|---|---|---|
| 2 | theory | | | | | -1/2 | 0 | 1/2 | | | | |
| | computed | 2 | | | | -0.500013397 | | 0.500013397 | | | | **2.6794E-05** |
| | | 3 | | | | -0.500013397 | 0.000000000 | 0.500013397 | | | | **2.6794E-05** |
| 4 | theory | | | | 1/12 | -2/3 | 0 | 2/3 | -1/12 | | | |
| | computed | 4 | | | 0.083342157 | -0.666684313 | | 0.666684313 | -0.083342157 | | | **5.2938E-05** |
| | | 5 | | | 0.083342315 | -0.666684942 | 0.000000938 | 0.666683691 | -0.083342002 | | | **5.3890E-05** |
| 6 | theory | | | -1/60 | 3/20 | -3/4 | 0 | 3/4 | -3/20 | 1/60 | | |
| | computed | 6 | | -0.016670578 | 0.150015647 | -0.750019558 | | 0.750019560 | -0.150015649 | 0.016670579 | | **7.8238E-05** |
| | | 7 | | -0.016669883 | 0.150011429 | -0.750008903 | -0.000014356 | 0.750030440 | -0.150020047 | 0.016671320 | | **9.3045E-05** |
| 8 | theory | | 1/280 | -4/105 | 1/5 | -4/5 | 0 | 4/5 | -1/5 | 4/105 | -1/280 | |
| | computed | 8 | 0.003565704 | -0.038061207 | 0.199921330 | -0.799922070 | | 0.799923558 | -0.199924308 | 0.038063121 | -0.003566130 | **3.8590E-04** |
| | | 9 | 0.003581062 | -0.038153753 | 0.200139250 | -0.800146107 | 0.000015479 | 0.800122711 | -0.200129608 | 0.038151977 | -0.003581011 | **6.8762E-04** |

*Table 6. Results of evolving coefficients of forward finite difference approximations of first derivative.*

| theoretical order of accuracy | | vector length | f(x) | f(x+h) | f(x+2h) | f(x+3h) | f(x+4h) | f(x+5h) | f(x+6h) | sum of absolute errors |
|---|---|---|---|---|---|---|---|---|---|---|
| 1 | theory | | -1 | 1 | | | | | | |
| | computed | 2 | -0.999965187 | 0.999992406 | | | | | | **4.2406E-05** |
| 2 | theory | | -3/2 | 2 | -1/2 | | | | | |
| | computed | 3 | -1.499934810 | 1.999923208 | -0.499988397 | | | | | **1.5359E-04** |
| 3 | theory | | -11/6 | 3 | -3/2 | 1/3 | | | | |
| | computed | 4 | -1.833239315 | 2.999797978 | -1.499878000 | 0.333319340 | | | | **4.3203E-04** |
| 4 | theory | | -25/12 | 4 | -3 | 4/3 | -1/4 | | | |
| | computed | 5 | -2.083211809 | 3.999619798 | -2.999588512 | 1.333164876 | -0.249984353 | | | **1.0973E-03** |
| 5 | theory | | -137/60 | 5 | -5 | 10/3 | -5/4 | 1/5 | | |
| | computed | 6 | -2.283198253 | 4.999457198 | -4.999179543 | 3.332777993 | -1.249854882 | 0.199997488 | | **2.2013E-03** |
| 6 | theory | | -49/20 | 6 | -15/2 | 20/3 | -15/4 | 6/5 | -1/6 | |
| | computed | 7 | -2.449833503 | 5.999156723 | -7.498280965 | 6.664893406 | -3.749059104 | 1.199779258 | -0.166655814 | **5.6746E-03** |

*Table 7. Results of evolving coefficients of two abnormal finite difference approximations of first derivative.*

| theoretical order of accuracy | | vector length | f(x-3h) | f(x-h) | f(x+h) | f(x+3h) | sum of absolute errors |
|---|---|---|---|---|---|---|---|
| 4 | theory | | 1/48 | -27/48 | 27/48 | -1/48 | |
| | computed | 4 | 0.020838333 | -0.562514996 | 0.562514996 | -0.020838333 | **3.9991E-05** |

| theoretical order of accuracy | | vector length | f(x-3h) | f(x-2h) | f(x-h) | f(x+h) | sum of absolute errors |
|---|---|---|---|---|---|---|---|
| 4 | theory | | 1/8 | -1/3 | -1/4 | 11/24 | |
| | computed | 4 | 0.124974036 | -0.333219716 | -0.250118502 | 0.458364183 | **2.8893E-04** |

*Table 8. Results of evolving coefficients of Adams-Bashforth schemes.*

| theoretical order of accuracy | | vector length | $f(t_{n-1}, y_{n-1})$ | $f(t_{n-1}, y_{n-1})$ | $f(t_{n-1}, y_{n-1})$ | $f(t_{n-1}, y_{n-1})$ | $f(t_{n-1}, y_{n-1})$ | sum of absolute errors |
|---|---|---|---|---|---|---|---|---|
| 1 | theory | | 1 | | | | | |
| | computed | 1 | 0.999998732 | | | | | **1.2676E-06** |
| 2 | theory | | 3/2 | -1/2 | | | | |
| | computed | 2 | 1.499995277 | -0.499995637 | | | | **9.0863E-06** |
| 3 | theory | | 23/12 | -4/3 | 5/12 | | | |
| | computed | 3 | 1.916658528 | -1.333318002 | 0.416659474 | | | **3.0663E-05** |
| 4 | theory | | 55/24 | -59/24 | 37/24 | -3/8 | | |
| | computed | 4 | 2.291653266 | -2.458294924 | 1.541630049 | -0.374988391 | | **1.0004E-04** |
| 5 | theory | | 1901/720 | -1387/360 | 109/30 | -637/360 | 251/720 | |
| | computed | 5 | 2.640277652 | -3.852780294 | 3.633341636 | -1.769452999 | 0.348614005 | **2.2393E-05** |

*Table 9. Results of evolving coefficients of Runge-Kutta schemes using fitness based on order conditions. The given numerical values are the most accuracte computed coefficient sets of accuracy order 3, 4 and 5, respectively.*

**3-stage**

| | | | | **best evolved scheme** | | |
|---|---|---|---|---|---|---|
| $c_1$ | | | | | | |
| $c_2$ | $a_{21}$ | | | 0.588205371365611 | | |
| $c_3$ | $a_{31}$ | $a_{32}$ | | -0.117042030825954 | 0.865356722666391 | |
| | $w_1$ | $w_2$ | $w_3$ | 0.239084361012680 | 0.433481022213682 | 0.327434616773638 |

**4-stage**

| | | | | | **best evolved scheme** | | | |
|---|---|---|---|---|---|---|---|---|
| $c_1$ | | | | | | | | |
| $c_2$ | $a_{21}$ | | | | 0.446027096189541 | | | |
| $c_3$ | $a_{31}$ | $a_{32}$ | | | -0.253232894933462 | 0.837303080381472 | | |
| $c_4$ | $a_{41}$ | $a_{42}$ | $a_{43}$ | | 0.284580085103288 | 0.018557477374513 | 0.696862437522200 | |
| | $w_1$ | $w_2$ | $w_3$ | $w_4$ | 0.160860757268920 | 0.410796107210609 | 0.268240761883735 | 0.160102373636736 |

**6-stage**

| | | | | | | |
|---|---|---|---|---|---|---|
| $c_1$ | | | | | | |
| $c_2$ | $a_{21}$ | | | | | |
| $c_3$ | $a_{31}$ | $a_{32}$ | | | | |
| $c_4$ | $a_{41}$ | $a_{42}$ | $a_{43}$ | | | |
| $c_5$ | $a_{51}$ | $a_{52}$ | $a_{53}$ | $a_{54}$ | | |
| $c_6$ | $a_{61}$ | $a_{62}$ | $a_{63}$ | $a_{64}$ | $a_{65}$ | |
| | $w_1$ | $w_2$ | $w_3$ | $w_4$ | $w_5$ | $w_6$ |

**best evolved scheme**

| | | | | | |
|---|---|---|---|---|---|
| 0.142950591304828 | | | | | |
| 0.737459236646687 | -0.504634588009118 | | | | |
| 0.314129433383799 | -0.330273585672633 | 0.467497950578092 | | | |
| -0.183250006068950 | 1.499638222192340 | -1.622659422172800 | 1.058063997380150 | | |
| -0.139352972771695 | -1.278258776673380 | 3.335534496262670 | -1.848343730854510 | 0.930420984036919 | |
| 0.008109845927407 | 0.365341829971006 | -0.104294398783786 | 0.327711908497619 | 0.318235531221308 | 0.084895283166445 |

*Table 10. Errors of best computed Runge-Kutta schemes*

| | **3-stage** | **4-stage** | **6-stage** |
|---|---|---|---|
| sum of absolute errors | $5.551 \cdot 10^{-17}$ | $2.637 \cdot 10^{-16}$ | $3.038 \cdot 10^{-14}$ |